\pdfoutput=1

\documentclass[11pt]{article}

\usepackage{emnlp2021}

\usepackage{times}
\usepackage{latexsym}

\usepackage[T1]{fontenc}

\usepackage[utf8]{inputenc}

\usepackage{microtype}
\usepackage{graphicx}
\usepackage{subfigure}
\usepackage{booktabs} 
\usepackage{url}
\usepackage{booktabs}

\usepackage{multirow}
\usepackage{colortbl}
\usepackage{color}
\usepackage{soul}
\usepackage{float}
\usepackage{graphics}
\usepackage{natbib}
\usepackage{enumitem}

\usepackage{amsmath,amsfonts,bm}

\def\eqref#1{equation~\ref{#1}}

\def\1{\bm{1}}

\def\vc{{\bm{c}}}

\def\vs{{\bm{s}}}

\def\vu{{\bm{u}}}
\def\vv{{\bm{v}}}

\def\mM{{\bm{M}}}

\DeclareMathAlphabet{\mathsfit}{\encodingdefault}{\sfdefault}{m}{sl}
\SetMathAlphabet{\mathsfit}{bold}{\encodingdefault}{\sfdefault}{bx}{n}

\def\gL{{\mathcal{L}}}

\newcommand{\R}{\mathbb{R}}

\usepackage{cleveref}
\Crefname{equation}{Eq.}{Eqns.}
\Crefname{figure}{Fig.}{Figs.}
\Crefname{theorem}{Thm.}{Thms.}

\title{Universal Sentence Representation Learning with Conditional Masked Language Model}

\author{Ziyi Yang$^1$\thanks{$\;\;$Work done during internship at Google Research.}$\;$, Yinfei Yang$^2$, Daniel Cer$^2$, Jax Law$^2$, Eric Darve$^1$ \\
$^1$Stanford University\\
\texttt{\{ziyi.yang,darve\}@stanford.edu}\\
$^2$Google Research\\
\texttt{\{yinfeiy,cer,jaxlaw\}@google.com}}

\begin{document}
\maketitle
\begin{abstract}
This paper presents a novel training method, Conditional Masked Language Modeling (CMLM), to effectively learn sentence representations on large scale unlabeled corpora. CMLM integrates sentence representation learning into MLM training by conditioning on the encoded vectors of adjacent sentences. Our English CMLM model achieves state-of-the-art performance on SentEval~\citep{senteval}, even outperforming models learned using supervised signals. As a fully unsupervised learning method, CMLM can be conveniently extended to a broad range of languages and domains. We find that a multilingual CMLM model co-trained with bitext retrieval~(BR) and natural language inference~(NLI) tasks outperforms the previous state-of-the-art multilingual models by a large margin, e.g. $10\%$ improvement upon baseline models on cross-lingual semantic search. We explore the same language bias of the learned representations, and propose a simple, post-training and model agnostic approach to remove the language identifying information from the representation while still retaining sentence semantics.
\end{abstract}

\section{Introduction}
Sentence embeddings map sentences into a vector space. The vectors capture rich semantic information that can be used to measure semantic textual similarity~(STS) between sentences or train classifiers for a broad range of downstream tasks~\citep{infersent,subramanian2018learning,qk,use,sentbert,muse,gem}.
State-of-the-art models are usually trained on supervised tasks such as natural language inference~\citep{infersent}, or with semi-structured data like question-answer pairs~\citep{use} and translation pairs~\citep{subramanian2018learning,muse}.
However, labeled and semi-structured data are difficult and expensive to obtain, making it hard to cover many domains and languages.
Conversely, recent efforts to improve language models include the development of masked language model (MLM) pre-training from large scale unlabeled corpora \citep{bert,albert,roberta}. 
While internal MLM model representations are helpful when fine-tuning on downstream tasks, they do not directly produce good sentence representations,
without further supervised \citep{sentbert} or semi-structured \citep{labse} fine-tuning.

\begin{figure*}[t]
  \centering
  \includegraphics[width=0.75\linewidth]{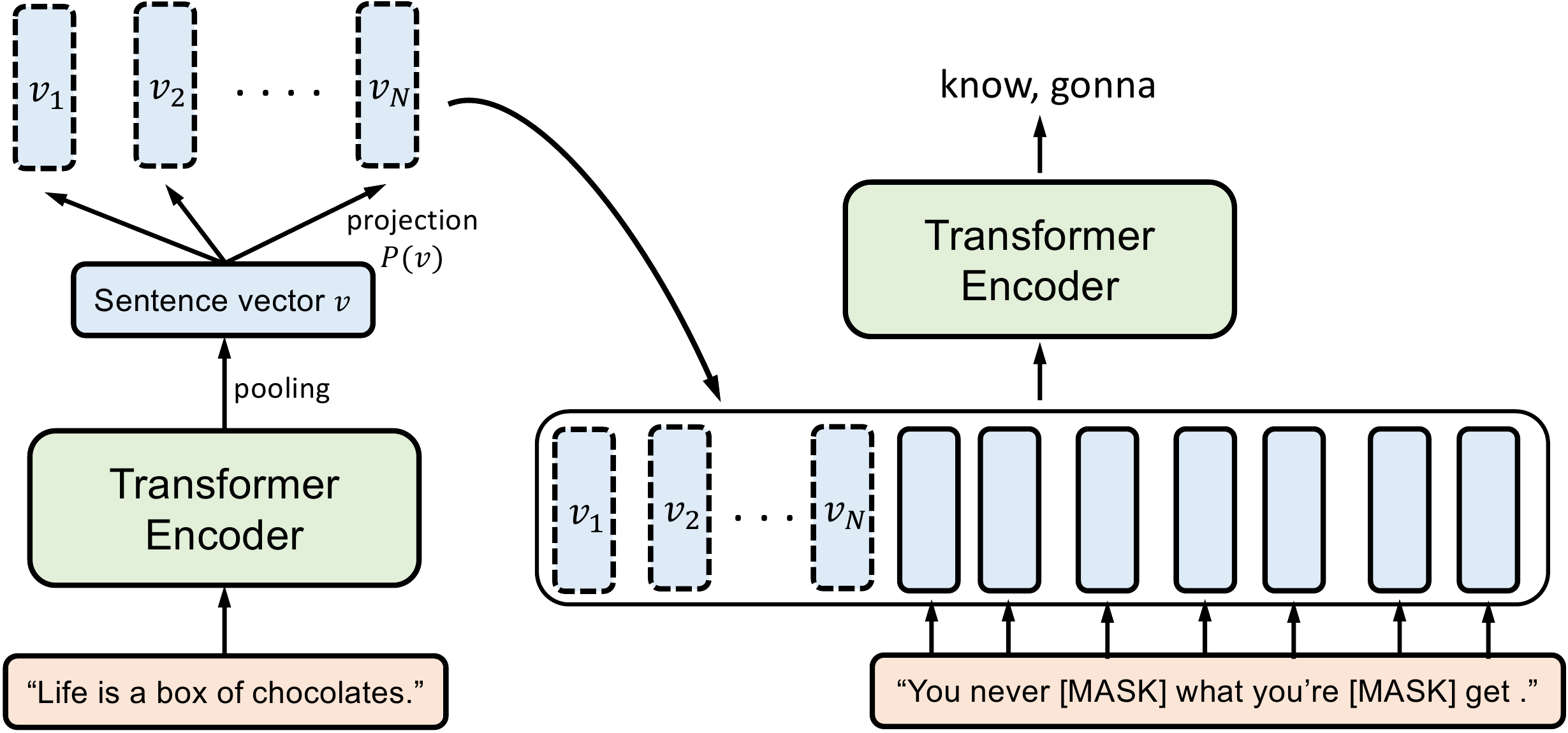}
  \caption{The architecture of Conditional Masked Language Modeling (CMLM).}
  \label{fig:cmlm}
\vskip -0.1in
\end{figure*}

In this paper, we explore an unsupervised approach, called Conditional Masked Language Modeling (CMLM), to effectively learn sentence representations from large scale unlabeled corpora. The CMLM model architecture is illustrated in \Cref{fig:cmlm}, which integrates sentence representation learning into MLM training by conditioning on sentence level representations produced by adjacent sentences. The model therefore needs to learn effective sentence representations in order to perform good MLM.
Since CMLM is fully unsupervised, it can be easily extended to new languages. We explore CMLM for both English and multilingual sentence embeddings for 100+ languages. 
Our English CMLM model achieves state-of-the-art performance on SentEval~\citep{senteval}, even outperforming models learned using (semi-)supervised signals. Moreover, models training on the English Amazon review data using our multilingual vectors exhibit strong multilingual transfer performance on translations of the Amazon review evaluation data to French, German and Japanese, outperforming existing multilingual sentence embedding models by $>5\%$ for non-English languages and by $> 2\%$ on English. 

We further extend the multilingual CMLM to co-train with parallel text (bitext) retrieval task, and finetune with cross-lingual natural language inference (NLI) data, inspired by the success of prior work on multitask sentence representation learning~\citep{subramanian2018learning,muse,distill-mling} and NLI learning~\citep{infersent,sentbert}. We achieve performance $3.6\%$ better than the previous state-of-the-art multilingual sentence representation model (see details in \Cref{sec:multitask}). On cross-lingual semantic search task, our model outperforms baseline models by $10\%$ on average over 36 languages.
Language agnostic representations require semantically similar cross-lingual pairs to be closer in representation space than unrelated same-language pairs~\citep{lareqa}. While we find our original sentence embeddings do have a bias for same language sentences, we discover that removing the first few principal components of the embeddings eliminates the self language bias.

The rest of the paper is organized as follows. \Cref{sec:cmlm} describes the architecture for CMLM unsupervised learning. In \Cref{sec:en_cmlm} we present CMLM trained on English data and evaluation results on SentEval. In \Cref{sec:mling_cmlm} we apply CMLM to learn sentence multilingual sentence representations. Multitask training strategies on how to effectively combining CMLM, bitext retrieval and cross-lingual NLI 
finetuning are explored. In \Cref{sec:analysis}, we investigate self language bias in multilingual representations and propose a simple but effective approach to eliminate it. The pre-trained models
are released at \url{https://tfhub.dev/s?q=universal-sentence-encoder-cmlm}.

\section{Conditional Masked Language Modeling}
\label{sec:cmlm}

We introduce Conditional Masked Language Modeling (CMLM) as a novel architecture for combining next sentence prediction with MLM training. By ``conditional'', we mean the MLM task for one sentence depends on the encoded sentence level representation of the adjacent sentence. This builds on prior work on next sentence prediction that has been widely used for learning sentence level representations \citep{skipthought,qk,use,muse}, but has thus far produced poor quality sentence embeddings within BERT based models using MLM loss~\citep{sentbert}. 

While existing MLMs like BERT include next sentence prediction tasks, they do so without any inductive bias to try to encode the meaning of a sentence within a single embedding vector. We introduce a strong inductive bias for learning sentence embeddings by structuring the task as follows. Given a pair of ordered sentences, the first sentence is fed to an encoder that produces a sentence level embedding. The embedding is then provided to an encoder that conditions on the sentence embedding in order to better perform MLM prediction over the second sentence. This is notably similar to Skip-Thought \citep{skipthought}, but replaces the generation of the complete second sentence with the MLM denoising objective. It is also similar to T5's MLM inspired unsupervised encode-decoder objective \citep{2019t5}, with the second encoder acting as a sort of decoder given the representation produced for the first sentence. Our method critically differs from T5's in that a sentence embedding bottleneck is used to pass information between two model components and in that the task involves denoising a second sentence when conditioning on the first rather than denoising a single text stream.

\Cref{fig:cmlm} illustrates the architecture of our model. The first sentence $s_1$ is tokenized and input to a transformer encoder and a sentence vector $\vv\in \R^{d}$ is computed from the sequence outputs by average pooling.\footnote{One can equivalently choose other pooling methods, such as max pooling or use the vector output corresponding to a special token position such as the [CLS] token.} The sentence vector $\vv$ is then projected into $N$ spaces with one of the projections being the identity mapping, i.e. $\vv_p = P(\vv) \in \R^{d\times N}$. Here we use a three-layer MLP as the projection $P(\cdot)$. Details of $P(\cdot)$ are available in the supplementary material. One motivation for the projections of $s_1$ is that MLM of $s_2$ then can attend to various representations of $s_1$ instead of only 1. In \Cref{sec:ab_std}, we explore various different configurations of CMLM, including the number of projection spaces $N$.

The second sentence $s_2$ is then masked following the procedure described in the original BERT paper, including random replacement and the use of unchanged tokens. The second encoder shares the same weights with the encoder used to embed $s_1$ \footnote{The dual-encoder sharing encoder weights for different inputs can be also referred as ``siamese encoder''}. Tokens in the masked $s_2$ are first converted into token vectors. The masked language modeling of $s_2$ depends on $s_1$ such that the process involves cross-attention between $s_2$ token vectors and $\vv_p$. In practice, this is implemented by concatenating token embeddings of $s_2$ with $\vv_p$\footnote{Representation concatenation has been used in previous work for enabling cross attention between \textit{global} vectors and \textit{local} token embeddings to help the representations learning of long/structured inputs~\cite{ainslie-etal-2020-etc,bigbird}.}.
Other implementations are also experimented (see \Cref{sec:ab_std}) and we empirically find concatenation works the best. The concatenated representations are then provided to the transformer encoder to predict the masked tokens in $s_2$.

At inference time, we keep the first encoding module and discard the subsequent MLM prediction. Similar to skip-thought, CMLM trains the encoder to produce sentence embeddings useful for predicting material in the adjunct sentences. CMLM adapts this existing idea to MLM training. Appending multiple projections performs well due to fine-grained attention between tokens and the different views of the sentence embeddings. Note that CMLM differs from SkipThought in the following aspects: (a) SkipThought relies on an extra decoder network while CMLM only has the encoder. (b) SkipThought predicts the entire sentence while CMLM predicts masked tokens only so the predictions can be done in parallel. These two differences make CMLM more efficient to train than SkipThought.

\section{Learning English Sentence Representations with CMLM}
\label{sec:en_cmlm}

For training English sentence encoders with CMLM, we use three Common Crawl dumps. The data are filtered by a classifier which detects whether a sentence belongs to the main content of the web page or not.
We use WordPiece tokenization and the vocabulary is the same as public English uncased BERT. In order to enable the model to learn bidirectional information, for two consecutive sequences $s_1$ and $s_2$, we swap their order for $50\%$ of the time. This order-swapping process echos with the preceding and succeeding sentences prediction in Skip-Thought \citep{skipthought}. The length of $s_1$ and $s_2$ are set to be 256 tokens (the maximum length). The number of masked tokens in $s_2$ are 80 (31.3\%), moderately higher than classical BERT. This change in the ratio of masked tokens is to make the task more challenging, due to the fact that in CMLM, language modeling has access to extra information from adjacent sentences. We train with batch size of 2048 for 1 million steps. The optimizer is LAMB \citep{lamb} with learning rate of $10^{-3}$, $\beta_1 = 0.9$, $\beta_2 = 0.999$, warm-up in the first 10,000 steps and linear decay afterwards. We explore two transformer configurations same as in the original BERT paper, i.e., base and large. The number of projections $N$ is 15 by experimenting with multiple choices.


\subsection{Evaluation}
We evaluate the sentence representations on the following tasks: (1) classification: MR (movie reviews \citet{mr}), binary SST (sentiment analysis, \citet{sst}), TREC (question-type, \citet{trec}), CR (product reviews, \citet{cr}), SUBJ (subjectivity/objectivity, \citet{subj}). (2) Entailment: SICK dataset for entailment (SICK-E, \citet{sick}). The evaluation is done using SentEval \citep{senteval} which is a prevailing evaluation toolkit for sentence embeddings.  The classifier for the downstream is logistic regression. For each task, the encoder and embeddings are fixed and only downstream neural structures are trained.

The baseline sentence embedding models include SkipThought \citep{skipthought}, InferSent \citep{infersent}, USE \citep{use}, QuickThought \citep{qk} and English BERT using standard pre-trained models from TensorFlow Hub website \citep{bert}, XLNet \citep{xlnet}, RoBERTa \citep{roberta}, SBert \citep{sentbert}. To evaluate the possible improvements coming from training data and processes, we train standard BERT models (English BERT base/large (CC)) on the same Common Crawl Corpora that CMLM is trained on. Similarly, we also train QuickThought, a competitive unsupervised sentence representations learning model, on the same Common Crawl Corpora (denoted as ``QuickThought (CC)''). To further address the possible advantage from using Transformer encoder, we use a Transformer encoder as the sentence encoder in QuickThought (CC). The representations for BERT are computed by averaging the sequence outputs (we also explore options including [CLS] vector and max pooling and the results are available in the appendix).

\subsection{Results}

Evaluation results are presented in \Cref{tab:en}. The numbers are averaged over 5 runs and the performance variances are provided in the appendix. CMLM outperforms existing models overall, besting MLM (both English BERT and English BERT (CC)) using both base and large configurations. The closest competing model is SBERT, which uses supervised NLI data rather than a purely unsupervised approach. Interestingly, CMLM outperforms SBERT on the SICK-E NLI task even the later model is trained with a NLI task. We also evaluate on Semantic Textual Similarity (STS) datasets. As shown in \Cref{tab:sts}, CMLM exhibits competitive performance compared with BERT and GloVe. One interesting observation is that CMLM base significantly outperforms other baselines (including CMLM large) on the STS Benchmark dataset. 

\begin{table*}[htb]
\small
\centering
\resizebox{\textwidth}{!}{\begin{tabular}[c]{l|cccccccccccc}
\toprule
  \textbf{Model} & \textbf{MR}  & \textbf{CR}   & \textbf{SUBJ} & \textbf{MPQA} & \textbf{SST}  & \textbf{TREC} &  \textbf{MRPC} & \textbf{SICK-E} & \textbf{SICK-R} & \textbf{Avg.}\\ \midrule
SkipThought & 76.5 & 80.1 & 93.6 & 87.1 & 82.0 & \textbf{92.2} & 73.0 & 82.3 & 85.8 & 83.8 \\
InferSent & 81.6 & \textbf{86.5} & 92.5 & \textbf{90.4} & 84.2 & 88.2 & \textbf{75.8} & \textbf{84.3} & \textbf{86.4} & \textbf{85.5} \\
USE & 80.1 & 85.2 & 94.0 & 86.7 & 86.4 & 93.2 & 70.1 & 82.4 & 85.9 & 84.9 \\
QuickThought (CC) & 75.7 & 81.9 & \textbf{94.3} & 84.7 & 79.7 & 83.0 & 70.4 & 75.0 & 78.5 & 80.4 \\
XLNet & \textbf{83.6} & 82.1 & 90.8 & 89.0 & \textbf{89.0} & 90.4 & 70.1 & 82.1 & 78.4 & 83.9 \\
\midrule
\multicolumn{10}{c}{\textit{BERT-based models}} \\
\midrule
English BERT base & 81.6 & 87.4 & 95.2 & 87.8 & 85.8 & 90.6 & 71.1 & 79.3 & 80.5 & 84.3\\
English BERT base (CC) & 82.5 & 88.5 & 95.6 & 87.3 & 88.0 & \textbf{91.4} & \textbf{72.0} & 79.3 & 79.0 & 84.6\\
SBERT (NLI, base) & \textbf{83.6} & 89.4 & 94.4 & \textbf{89.9} & \textbf{88.9} & 89.6 & 76.0 & 79.9 & 80.6 & 85.8 \\  
\textbf{CMLM base (ours)} & \textbf{83.6} & \textbf{89.9} & \textbf{96.2} & 89.3 & 88.5 & 91.0 & 69.7 & \textbf{82.3} & \textbf{83.4} & \textbf{86.0} \\\midrule
English BERT large & 84.3 & 88.9 & 95.7 & 86.8 & 88.9 & 91.4 & 71.8 & 75.7 & 77.0 & 84.5 \\
English BERT large (CC) & 85.4 & 89.0 & 95.7 & 86.9 & 90.5 & 91.2 & 75.5 & 74.3 & 77.0 & 85.0 \\
RoBERTa (large) & 85.2 & \textbf{90.6} & \textbf{97.0} & 90.0 & 89.5 & 93.6 & 74.2 & 75.1 & 78.9 & 86.0 \\ 
SBERT (NLI, large) & 84.8 & 90.0 & 94.5 & \textbf{90.3} & 90.7 & 87.4 & \textbf{76.0} & 74.9 & 84.2 & 85.9 \\
\textbf{CMLM large (ours)} & \textbf{85.6} & 89.1 & 96.6 & 89.3 & \textbf{91.4} & \textbf{92.4} & 70.0 & \textbf{82.2} & \textbf{84.5} & \textbf{86.8}\\ \toprule
\end{tabular}}
\vskip -0.1in
\caption{Transfer learning test set results on SentEval for English models. Baseline models include BERT-based (BERT, RoBERTA and SBERT) and non-BERT models (XLNet, SkipThought, InferSent and USE).}
\label{tab:en}
\end{table*}

\begin{table*}[htb]
\small
\centering
\resizebox{\textwidth}{!}{\begin{tabular}[c]{l|ccccccccc}
\toprule
  \textbf{Model} & \textbf{STS12}  & \textbf{STS13}   & \textbf{STS14} & \textbf{STS15} & \textbf{STS16}  & \textbf{STSB} &  \textbf{SICK-R} & \textbf{Avg.}\\ \midrule
Avg. GloVe embeddings & 55.14 & \textbf{70.66} & 59.73 & 68.25 & 63.66 & 58.02 & 53.76 & 61.32 \\
BERT Mean embeddings & 38.78 & 57.98 & 57.98 & 63.15 & 61.06 & 46.35 & 58.40 & 54.81 \\
BERT CLS-vector & 20.16 & 30.01 & 20.09 & 36.88 & 38.08 & 16.50 & 42.63 & 29.19 \\
\textbf{CMLM base (ours)} & 58.20 & 61.07 & 61.67 & 73.32 & 74.88 & \textbf{76.60} & 64.80 & \textbf{67.22} \\
\textbf{CMLM large (ours)} & \textbf{59.02} & 61.68 & \textbf{62.80} & \textbf{74.16} & \textbf{75.64} & 69.39 & \textbf{66.56} & 67.03 \\ \bottomrule
\end{tabular}}
\vskip -0.1in
\caption{Spearman rank correlation on Semantic Textual Similarity (STS) datasets. SICK-R is zero-shot evaluation by directly computing the cosine similarity of sentence embeddings, without training the task-specific neural network.}
\label{tab:sts}
\end{table*}

\section{Learning Multilingual Sentence Representations with CMLM}
\label{sec:mling_cmlm}

As a fully unsupervised method, CMLM can be conveniently extended to multilingual modeling even for less well resourced languages. Learning good multilingual sentence representations is more challenging than monolingual ones, especially when attempting to capture the semantic alignment between different languages. As CMLM does not explicitly address cross-lingual alignment, we explore several modeling approaches besides CMLM: (1) Co-training CMLM with a bitext retrieval task; (2) Fine-tuning with cross-lingual NLI data.

\subsection{Multilingual CMLM}
We follow the same configuration used to learn English sentence representations with CMLM, but extend the training data to include more languages. 
Results below will show that CMLM again exhibits competitive performance as a general technique to learn from large scale unlabeled corpora.

\subsection{Multitask Training with CMLM and Bitext Retrieval}
\label{sec:multitask}
Besides the monolingual pretraining data, we collect a dataset of bilingual translation pairs \{($s_{i}$, $t_{i}$)\} using a bitext mining system~\citep{labse}. The source sentences $\{s_{i}\}$ are in English and the target sentences $\{t_{i}\}$ covers over 100 languages. We build a retrieval task with the translation parallel data, identifying the corresponding translation of the input sentence from candidates in the same batch. Concretely, incorporating Additive Margin Softmax~\citep{ijcai2019-746}, we compute the bitext retrieval loss $\gL^{s}_{br}$ for the source sentences as:

$$\gL^s_{br}=-\frac{1}{B} \sum_{i=1}^{B} \frac{e^{\phi\left(s_{i}, t_{i}\right)-m}}{e^{\phi\left(s_{i}, t_{i}\right)-m}+\sum_{j=1, j \neq i}^{B} e^{\phi\left(s_{i}, t_{j}\right)}}$$


Above $\phi(s_{i}, t_{j})$ denotes the the inner products of sentence vectors of $s_i$ and $t_j$ (embedded by the transformer encoder); $m$ and $B$ denotes the additive margin and the batch size respectively. Note the way to generate sentence embeddings is the same as in CMLM. We can compute the bitext retrieval loss for the target sentences $\gL^{t}_{br}$ by normalizing over source sentences, rather than target sentences, in the denominator.\footnote{i.e., by swapping the $i$ and $j$ subscripts in the last term of the denominator.} The final bitext retrieval loss $\gL_{br}$ is given as $\gL_{br} = \gL^s_{br} + \gL^t_{br}$.

There are several ways to incorporate the monolingual CMLM task and bitext retrieval (BR). We explore the following multistage and multitask pretraining strategies: 
\begin{itemize}[noitemsep,topsep=0pt]
    \item[S1.] CMLM+BR: Train with CMLM and BR from the start;
    \item[S2.] CMLM~$\rightarrow$~BR: Train with CMLM in the first stage and then train with on BR;
    \item[S3.] CMLM~$\rightarrow$~CMLM+BR: Train with only CMLM in the first stage and then with both tasks.
\end{itemize}
When training with both CMLM and BR, the optimization loss is a weighted sum of the language modeling and the retrieval loss $\gL_{br}$, i.e. $\gL = \gL_{CMLM} + \alpha \gL_{br}$. We empirically find $\alpha=0.2$ works well. As shown in \Cref{tab:xeval}, S3 is found to be the most effective. Unless otherwise denoted, our models trained with CMLM and BR follow S3. We also discover that given a pre-trained transformer encoder, e.g. mBERT, we can improve the quality of sentence representations by finetuning the transformer encoder with CMLM and BR. As shown in \Cref{tab:xeval}, the improvements of f-mBERT (finetuned mBERT) upon mBERT are significant.

\begin{table*}[htbp]
\centering
\resizebox{\linewidth}{!}{
\begin{tabular}{lcccccccccccccccc}
\toprule
Model & \textbf{ar} & \textbf{bg} & \textbf{de} & \textbf{el} & \textbf{en} & \textbf{es} & \textbf{fr} & \textbf{hi} & \textbf{ru} & \textbf{sw} & \textbf{th} & \textbf{tr} & \textbf{ur} & \textbf{vi} & \textbf{zh} & \textbf{Avg.} \\ \midrule
mBERT & 76.3 & 76.1 & 77.7 & 76.1 & 80.1 & 78.5 & 78.7 & 75.6 & 77.3 & 70.5 & 73.6 & 75.7 & 74.2 & 78.8 & 78.7 & 76.5 \\ 
MLM (CC) & 79.2 & 79.1 & 81.7 & 79.9 & 84.4 & 82.1 & 82.2 & 79.2 & 81.2 & 70.3 & 76.9 & 79.0 & 74.3 & 81.3 & 81.0 & 79.4 \\ 
XLM-R & 78.1 & 78.0 & 76.2 & 78.2 & 82.8 & 81.2 & 80.4 & 77.2 & 80.2 & 71.0 & 77.5 & 79.7 & 76.7 & 80.3 & 80.8 & 78.5 \\ 
\textbf{CMLM} & \textbf{80.6} & \textbf{81.2} & \textbf{82.6} & \textbf{81.4} & \textbf{85.0} & \textbf{82.3} & \textbf{83.4} & \textbf{80.0} & \textbf{82.3} & \textbf{76.2} & \textbf{78.8} & \textbf{81.0} & \textbf{78.5} & \textbf{81.6} & \textbf{81.7} & \textbf{81.2} \\ \bottomrule
\end{tabular}}
\vskip -0.1in
\caption{Performance (accuracy) of multilingual models trained with monolingual data on XEVAL. Highest numbers are highlighted in bold.}
\label{tab:xeval_mono}
\end{table*}

\begin{table*}[]
\centering
\resizebox{\textwidth}{!}{
\begin{tabular}{lcccccccccccccccc}
\toprule
Model & \textbf{ar} & \textbf{bg} & \textbf{de} & \textbf{el} & \textbf{en} & \textbf{es} & \textbf{fr} & \textbf{hi} & \textbf{ru} & \textbf{sw} & \textbf{th} & \textbf{tr} & \textbf{ur} & \textbf{vi} & \textbf{zh} & \textbf{Avg.} \\ \midrule
LASER & 82.1 & 81.2 & 81.7 & 78.1 & 82.3 & 81.0 & 80.8 & 78.9 & 82.2 & 75.8 & 80.3 & 81.8 & 77.2 & 81.6 & 82.1 & 80.4 \\
mUSE  & 80.4 & -- & 82.2 & -- & 83.3 & 82.7 & 82.4 & -- & 82.3 & -- & 81.6 & 80.3 & -- & -- & 82.0 & 81.9 \\
\textbf{S1} & 78.3 & 78.9 & 79.3 & 78.1 & 81.0 & 78.7 & 79.5 & 78.0 & 79.0 & 76.6 & 77.8 & 78.6 & 77.7 & 79.0 & 78.6 & 78.6 \\ 
\textbf{S2} & 81.3 & 81.0 & 83.0 & 81.4 & 85.6 & 83.0 & 83.6 & 80.4 & 82.3 & 77.6 & 80.1 & 81.0 & 79.8 & 82.4 & 82.3 & 81.6 \\
\textbf{S3} & 82.6 & 83.0 & 84.0 & 81.8 & 85.8 & 84.2 & 84.6 & 81.7 & 84.0 & 79.3 & 81.2 & 82.7 & 81.2 & 83.0 & 83.0 & 82.8 \\
\textbf{S3+NLI} & \textbf{84.2} & \textbf{83.7} & \textbf{85.0} & \textbf{83.4} & \textbf{87.0} & \textbf{85.9} & \textbf{85.8} & \textbf{83.0} & \textbf{85.6} & \textbf{79.6} & \textbf{83.0} & \textbf{84.2} & \textbf{81.2} & \textbf{84.2} & \textbf{84.4} & \textbf{84.0} \\ 
\midrule
mBERT & 76.3 & 76.1 & 77.7 & 76.1 & 80.1 & 78.5 & 78.7 & 75.6 & 77.3 & 70.5 & 73.6 & 75.7 & 74.2 & 78.8 & 78.7 & 76.5 \\ 
\textbf{f-mBERT} & 77.2 & 78.5 & 79.7 & 76.7 & 81.4 & 80.0 & 80.3 & 77.2 & 79.1 & 73.3 & 76.1 & 77.1 & 76.9 & 79.8 & 80.4 & 78.3 \\ 
\bottomrule
\end{tabular}}
\vskip -0.1in
\caption{Performance (accuracy) of models trained with cross-lingual data on XEVAL. We test with multiple strategies for multitask pretraining: \textbf{[S1]}: CMLM$~\rightarrow$~BR; \textbf{[S2]}: CMLM+BR; \textbf{[S3]}: CMLM~$\rightarrow$~CMLM+BR. \textbf{[f-mBERT]} denotes finetuning mBERT with CMLM and BR.}
\label{tab:xeval}
\end{table*}

\subsection{Finetuning with Cross-lingual Natural Language Inference} 
Finetuning with NLI data has proved to be an effective method to improve the quality of embeddings for English models. We propose to leverage cross-lingual NLI finetuning in multilingual representations. Given a premise sentence $\vu$ in language $l_1$ and a hypothesis sentence $\vv$ in language $l_2$, we train a 3-way classifier on the concatenation of $[\vu, \vv, |\vu - \vv|, \vu*\vv]$. Weights of transformer encoders are also updated in the finetuning process. Different from previous work also using multilingual NLI data \citep{muse}, the premise $\vu$ and hypothesis $\vv$ are in \textbf{different} languages. The cross-lingual NLI data are generated by translating Multi-Genre NLI Corpus \citep{multinli} into 14 languages using Google Translate API.

\subsection{Configurations}

Monolingual training data for CMLM are generated from 3 versions of Common Crawl data in 113 languages. The data cleaning and filtering is the same as the English-only ones. A new cased vocabulary is built from the all
data sources using the WordPiece vocabulary generation library from Tensorflow Text. The language smoothing exponent from the vocab generation tool is set to 0.3, as the distribution of data size for each language is imbalanced. The final vocabulary size is 501,153. The number of projections $N$ is set to be $15$, the batch size $B$ is 2048 and the positive margin is $0.3$.  For CMLM only pretraining, the number of steps is 2 million. In multitask learning, for S1 and S3, the first stage is of 1.5 million and the second stage is of 1 million steps; for S2, number of training steps is 2 million. The transformer encoder uses the BERT base configuration. Initial learning rate and optimizer chosen are the same as the English models. Motivations for choosing such configurations, training details and potential limitations of CMLM are discussed in the appendix.

\begin{table*}[htb]
\centering
\resizebox{\linewidth}{!}{
\begin{tabular}{l|ccccccccccccccccccc}
\toprule
Lang. & af & ar & bg & bn & de & el & es & et & eu & fa & fi & fr & he & hi & hu & id & it & ja & \\
\midrule
mBERT  & 42.7 & 25.8 & 49.3 & 17 & 77.2 & 29.8 & 68.7 & 29.3 & 25.5 & 46.1 & 39 & 66.3 & 41.9 & 34.8 & 38.7 & 54.6 & 58.4 & 42 \\
MLM (CC) & 60.5 & 51.4 & 74.8 & 45   & 89.3 & 68.3 & 81.8 & 56.8 & 59.5 & 81.3 &  76.6 & 82.6 & 72.2 & 76.2 & 68.4 & 82.6 & 72.8 & 65.7 \\
XLM   & 43.2 & 18.2 & 40 & 13.5 & 66.2 & 25.6 & 58.4 & 24.8 & 17.1 & 32.2 & 32.2 & 54.5 & 32.1 & 26.5 & 30.1 & 45.9 & 56.5 & 40 \\
XLM-R  & 58.2 & 47.5 & 71.6 & 43 & 88.8 & 61.8 & 75.7 & 52.2 & 35.8 & 70.5 & 71.6 & 73.7 & 66.4 & 72.2 & 65.4 & 77 & 68.3 & 60.6 \\
LASER & 89.4 & \textbf{91.9} & 95.0 & 89.6 & \textbf{99.0} & 94.9 & 98.0 & \textbf{96.7} & \textbf{94.6} & 71.6 & \textbf{96.3} & 95.6 & 92.1 & 94.7 & 96.0 & 94.5 & \textbf{95.4} & 95.3 \\
\textbf{CMLM} & 62.0 & 53.2 & 75.0 & 45.1 & 89.9 & 69.9 & 82.7 & 59.2 & 61.6 & 83.7 &  77.1 & 83.5 & 73.1 & 76.7 & 70.3 & 83.0 & 73.5 & 67.2 \\
\textbf{CMLM+BR} & \textbf{96.3} & 90.6 & \textbf{95.4} & \textbf{91.2} & 97.7 & \textbf{95.4} & \textbf{98.1} & 95.6 & 92.0 & \textbf{95.6} & 95.9 & \textbf{96.1} & \textbf{92.8} & \textbf{97.6} & \textbf{96.5} & \textbf{95.6} & 94.2 & \textbf{95.6} \\
\textbf{CMLM+BR+NLI} & 90.5 & 83.6 & 92.6 & 86.4 & 97.6 & 91.6 & 95.5 & 82.6 & 76.3 & 90.7 & 88.9 & 93.5 & 86.8 & 94.6 & 89.6 & 91.7 & 90.4 & 88.4\\
\midrule
& jv & ka & kk & ko & ml & mr & nl & pt & ru & sw & ta & te & th & tl & tr & ur & vi & zh & Mean\\
\midrule
mBERT & 17.6 & 20.5 & 27.1 & 38.5 & 19.8 & 20.9 & 68 & 69.9 & 61.2 & 11.5 & 14.3 & 16.2 & 13.7 & 16 & 34.8 & 31.6 & 62 & 71.6 & 38.7 \\
MLM (CC) & 49.5 & 65.8 & 61.3 & 66.4 & 65.3 & 56.8 & 83.4 & 83.1 & 74.8 & 65.9 & 61.3 & 68.5 & 70.0 & 62.7 & 70.3 & 80.1 & 77.0 & 71.3 & 69.4 \\
XLM  & 22.4 & 22.9 & 17.9 & 25.5 & 20.1 & 13.9 & 59.6 & 63.9 & 44.8 & 12.6 & 20.2 & 12.4 & 31.8 & 14.8 & 26.2 & 18.1 & 47.1 & 42.2 & 32.6 \\
XLM-R  & 14.1 & 52.1 & 48.5 & 61.4 & 65.4 & 56.8 & 80.8 & 82.2 & 74.1 & 20.3 & 26.4 & 35.9 & 29.4 & 36.7 & 65.7 & 24.3 & 74.7 & 68.3 & 57.3 \\
LASER & 23.0 & 35.9 & 18.6 & 88.9 & 96.9 & 91.5 & 96.3 & 95.2 & 94.4 & 57.5 & 69.4 & 79.7 & 95.4 & 50.6 & 97.5 & 81.9 & 96.8 & 95.5 & 84.4 \\
\textbf{CMLM} & 51.8 & 65.5 & 62.7 & 67.2 & 65.8 & 57.0 & 83.8 & 83.6 & 75.5 & 66.6 & 61.7 & 68.8 & 70.3 & 63.5 & 70.5 & 80.3 & 77.4 & 71.7 & 70.3 \\
\textbf{CMLM+BR} & \textbf{83.4} & \textbf{94.9} & \textbf{88.6} & \textbf{92.4} & \textbf{98.9} & \textbf{94.5} & \textbf{97.3} & \textbf{95.3} & \textbf{94.9} & \textbf{87.0} & \textbf{91.2} & \textbf{97.9} & \textbf{96.6} & \textbf{95.3} & \textbf{98.6} & \textbf{94.4} & \textbf{97.5} & \textbf{95.6} & \textbf{94.7} \\
\textbf{CMLM+BR+NLI} & 66.9 & 88.1 & 80.3 & 85.6 & 94.9 & 90.7 & 93.2 & 92.3 & 91.7 & 76.7 & 88.6 & 92.8 & 94.7 & 82.0 & 94.3 & 84.7 & 94.3 & 93.1 & 88.8 \\
\bottomrule
\end{tabular}}
\vskip -0.1in
\caption{Tatoeba results (retrieval accuracy) for each language. Our model CMLM+BR achieves the best results on 30 out of 36 languages.}
\label{tab:tatoeba}
\end{table*}

\begin{table}[t!]
\centering
\resizebox{\columnwidth}{!}{
\begin{tabular}[c]{lcccc}
\toprule
Models & English & French & German & Japanese \\ \midrule
\multicolumn{5}{l}{\textit{Encoder parameters are frozen during finetuning}}\\
\midrule
\citet{eriguchi2018zero}  & 83.2 & 81.3 & - & -\\
MTDE en-fr & 87.4 & 82.3 & - & - \\
MTDE en-de & 87.1 & - & 81.0 & - \\
mBERT  & 80.0 & 73.1 & 70.4 & 71.7 \\
XLM-R & - & 85.3 & 81.5 & 82.5 \\
MLM (CC) & 84.6 & 84.9 & 84.3 & 82.1 \\
\textbf{CMLM}  & 88.4 & 88.2 & 87.5 & \textbf{83.7} \\
\textbf{CMLM+ BR}  & 88.3 & 87.2 & 86.4 & 83.2 \\
\textbf{CMLM+ BR + NLI} & \textbf{89.4} & \textbf{88.8} & \textbf{88.4} & 82.8\\
\midrule
\multicolumn{5}{l}{\textit{Encoder parameters are trained during finetuning}}\\
\midrule
mBERT  & 89.3 & 83.5 & 79.4 & 74.0 \\
MLM (CC) & 92.9 & 88.7 & 88.4 & 86.3 \\
\textbf{CMLM}  & 93.4 & 92.4 & 92.1 & \textbf{88.6} \\
\textbf{CMLM+ BR}  & 93.6 & \textbf{93.1} & 92.3 & 88.1 \\
\textbf{CMLM+ BR + NLI}  & \textbf{93.7} & 92.4 & \textbf{93.5} & 86.8 \\
\bottomrule
\end{tabular}}
\vskip -0.1in
\caption{Classification accuracy on the Amazon Reviews dataset.}
\label{tab:amazon}
\vskip -0.1in
\end{table}

\subsection{Evaluations}
\subsubsection{XEVAL: Multilingual Benchmarks for Sentence Representations Evaluation}

Evaluations in previous multilingual literature focused on the cross-lingual transfer learning ability from English to other languages. However, this evaluation protocol that treats English as the ``anchor'' does not equally assess the quality of non-English sentence representations with English ones. To address the issue, we prepare a new benchmark for multilingual sentence vectors, XEVAL, by translating SentEval (English) to other 14 languages with Google Translate API. The reliability of XEVAL is discussed in the appendix.

Results of models trained with monolingual data are shown in \Cref{tab:xeval_mono}. Baseline models include mBERT \citep{bert}, XLM-R \citep{xlm-r} and a transformer encoder trained with MLM on the same Common Crawl data (MLM(CC), again this is to control the effects of training data). The method to produce sentence representations for mBERT and XLM-R is chosen to be average pooling after exploring options including [CLS] representations and max pooling. The multilingual model CMLM trained on monolingual data outperform all baselines in all 15 languages.

Results of models trained with cross-lingual data are presented in \Cref{tab:xeval}. Baseline models for comparison include LASER (\citet{laser}, trained with parallel data) and multilingual USE (\citep{muse}, trained with cross-lingual NLI. Note it only supports 16 languages). Our model (S3) outperforms LASER in all 15 languages. Notably, finetuning with NLI in the cross-lingual way produces significant improvement (S3 + NLI v.s. S3). Multitask learning with CMLM and BR can also be used to increase the performance of pretrained encoders, e.g. mBERT. mBERT trained with CMLM and BR (f-mBERT) has a significant improvement upon mBERT.

\begin{table*}[t!]
\centering
\resizebox{\textwidth}{!}{
\begin{tabular}[c]{l|ccccccccccc}
\toprule
  \textbf{Model} & \textbf{MR}  & \textbf{CR}   & \textbf{SUBJ} & \textbf{MPQA} & \textbf{SST}  & \textbf{TREC} &  \textbf{MRPC} & \textbf{SICK-E} & \textbf{SICK-R} & \textbf{Avg.}\\ \midrule
$N=1$ & 82.3 & 89.7 & 95.8 & 88.8 & 87.6 & 90.4 & \textbf{71.5} & 80.8 & 83.4 & 85.5 \\
$N = 5$ & \textbf{83.7} & \textbf{90.0} & 95.5 & 89.0 & \textbf{89.4} & 86.6 & 69.5 & 79.3 & 81.7 & 85.0 \\
$N = 10$ & 83.4 & 89.0 & 96.1 & 88.9 & 88.2 & 90.2 & 68.5 & 79.7 & 81.5 & 84.9 \\
$N = 15$ & 83.6 & 89.9 & \textbf{96.2} & \textbf{89.3} & 88.5 & \textbf{91.0} & 69.7 & \textbf{82.3} & \textbf{83.4} & \textbf{86.0} \\
$N = 20$ & 81.1 & 89.5 & 95.8 & 88.9 & 85.9 & 89.8 & 69.7 & 80.2 & 85.0 & 85.1 \\
skip & 80.3 & 86.8 & 94.5 & 87.5 & 84.9 & 86.0 & 69.2 & 72.8 & 74.7 & 81.9 \\ 
proj & 82.6 & 89.7 & 96.0 & 87.3 & 87.5 & 89.2 & 70.5 & 81.7 & 83.8 & 85.4 \\
\bottomrule
\end{tabular}}
\vskip -0.1in
\caption{Ablation study of CMLM designs, including the number of projection spaces, architecture and sentence representations. The experiments are conducted on SentEval.}
\label{tab:abl}
\end{table*}

\subsubsection{Amazon Reviews}
We conduct a zero-shot transfer learning evaluation on Amazon reviews dataset \citep{amazon}. Following \citet{chidambaram}, the original dataset is converted to a classification benchmark by treating reviews with strictly more than 3 stars as positive and negative otherwise. We split 6000 English reviews in the original training set into 90\% for training and 10\% for development. The two-way classifier, upon the concatenation of $[\vu, \vv, |\vu - \vv|, \vu*\vv]$ (following works e.g. \citet{sentbert}), is trained on the English training set and then evaluated on English, French, German and Japanese test sets (each has 6000 examples). The same multilingual encoder and classifier are used for all the evaluations. We also experiment with whether freezing the encoder weights or not during training. As presented in \Cref{tab:amazon}, CMLM alone has already outperformed baseline models, including Multi-task Dual-Encoder (MTDE, \citet{chidambaram}), mBERT and XLM-R. Training with BR and cross-lingual NLI finetuning further boost the performance.

\subsection{Tatoeba: Semantic Search}
We test on Tatoeba dataset proposed in \citet{laser} to asses the ability of our models on capturing cross-lingual semantics. The task is to find the nearest neighbor for the query sentence in the other language. The experiments is conducted on the 36 languages as in XTREME \citep{xtreme}. The evaluation metric is retrieval accuracy. Results are presented in \Cref{tab:tatoeba}. Our model CMLM+BR outperforms all baseline models in 30 out of 36 languages and has the highest average performance. One interesting observation is that finetuning with NLI actually undermines the model performance on semantic search, in contrary with the significant improvements from CMLM+BR to CMLM+BR+NLI on XEVAL (\Cref{tab:xeval}). We speculate this is because unlike semantic search, NLI inference is not a linear process. Finetuning with NLI is not expected to help the linear retrieval by nearest neighbor search.

\begin{figure*}[t!]
  \centering
  \includegraphics[width=0.8\linewidth]{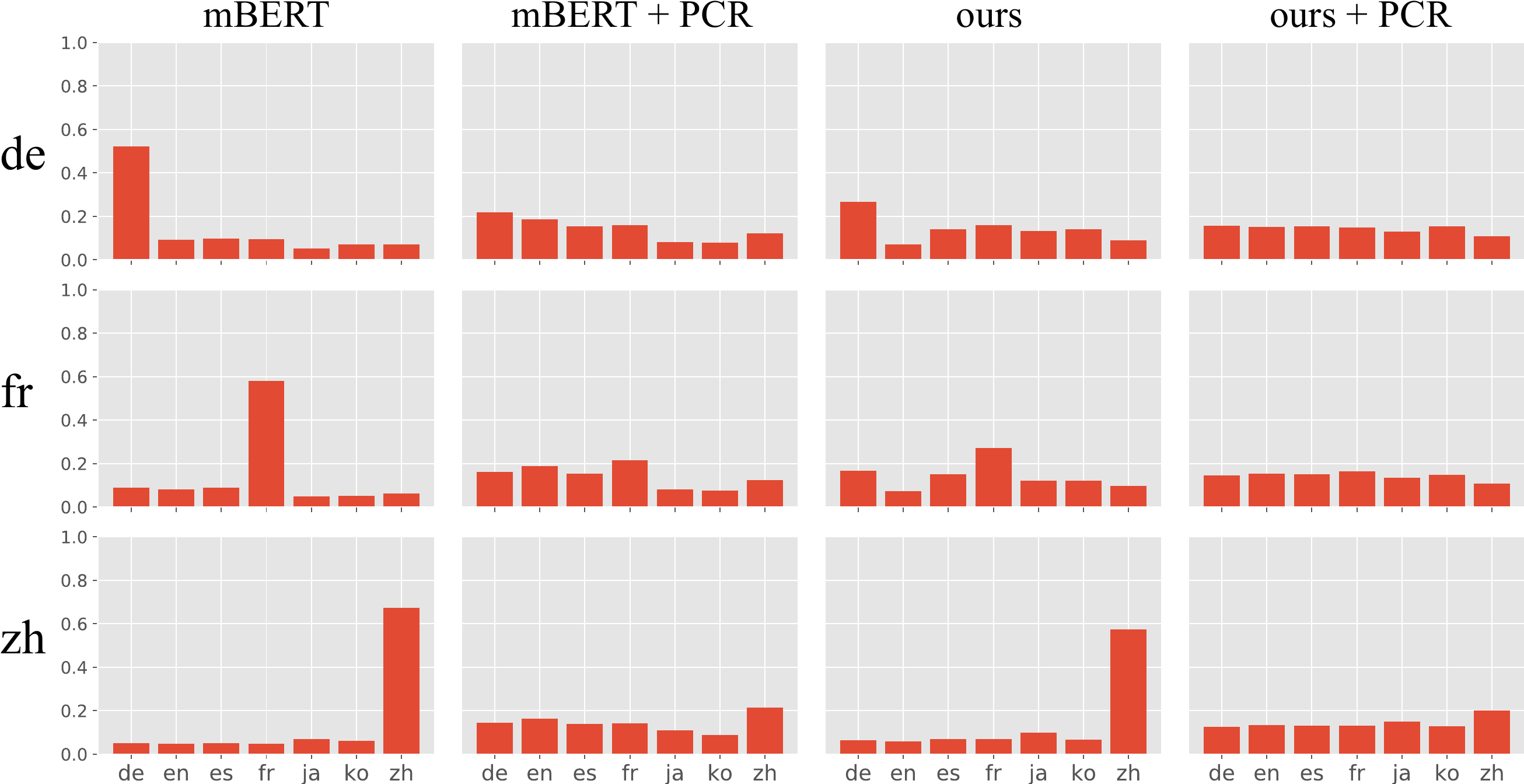}
  \vskip -0.1in
  \caption{Language distribution of retrieved sentences. The histogram values represent the percentage of sentences retrieved in a language. The first and third columns are mBERT and our models. Our model already in general has a more uniform distribution than mBERT. The second and fourth columns are mBERT and our model with PCR.}
  \label{fig:la}
\end{figure*}

\section{Analysis}
\label{sec:analysis}
\subsection{Ablation Study}
\label{sec:ab_std}

We explore different configurations of CMLM, including the number of projection spaces $N$ (\Cref{tab:abl}). Projecting the sentence vector into $N=15$ spaces produces highest overall performance. We also try a different CMLM architecture. Besides the concatenation with token embeddings of $s_2$ before input to the transformer encoder, the projected vectors are also concatenated with the sequence outputs of $\vs_2$ for the masked token prediction. This version of architecture is denoted as ``skip'' and the model performance is actually worse.

Note that the projected vector can also be used to produce the sentence representation $\vv_s$, e.g. using the average of projected vectors $\vv_s = \frac{1}{N}\sum_{i} \vv_p^{(i)}$ as the sentence embeddings. Recall $\vv_p^{(i)}$ is the $i$th projection. This version is denoted as ``proj'' in \Cref{tab:abl}. Sentence representations produced in this way still yield competitive performance, which further confirm the effectiveness of the projection.

\begin{table}[htbp]
\centering
\vskip -0.1in
\resizebox{\columnwidth}{!}{\begin{tabular}[c]{l|ccccccccccccccc}
\toprule
 & fra & cmn & spa & deu & rus & ita\\ \midrule
mBERT & \textbf{60.2} & 60.2 & \textbf{62.8} & 65.9 & 53.8 & 55.7 \\
mBERT + PCR & 59.9 & \textbf{64.3} & 61.7 & \textbf{67.5} & \textbf{57.4} & \textbf{56.2} \\ \midrule

ours & \textbf{96.1} & 95.6 & 98.1 & 97.7 & 94.9 & \textbf{94.2} \\
ours + PCR & 95.5 & \textbf{96.0} & \textbf{98.2} & \textbf{97.9} & \textbf{95.1} & 94.1 \\
\toprule
 & tur & por & hun & jpn & nld & Avg. \\
\midrule
mBERT & 32.4 & 62.4 & 31.9 & 39.0 & 56.2 & 52.8 \\
mBERT + PCR & \textbf{33.3} & \textbf{64.4} & \textbf{36.5} & \textbf{42.3} & \textbf{61.1} & \textbf{54.8} \\ \midrule
ours & \textbf{98.6} & 95.3 & 96.5 & \textbf{95.6} & \textbf{97.3}  & 96.3 \\
ours + PCR & 98.5 & \textbf{95.8} & \textbf{96.6} & 95.3 & 97.2 & \textbf{96.4} \\
\bottomrule
\end{tabular}}
\vskip -0.1in
\caption{Average retrieval accuracy on 11 languages of multilingual representations model with and without PCR on Tatoeba dataset.}
\label{tab:tatoeba_pcr}
\vskip -0.1in
\end{table}

\begin{figure*}[t!]
  \centering
  \includegraphics[width=\linewidth]{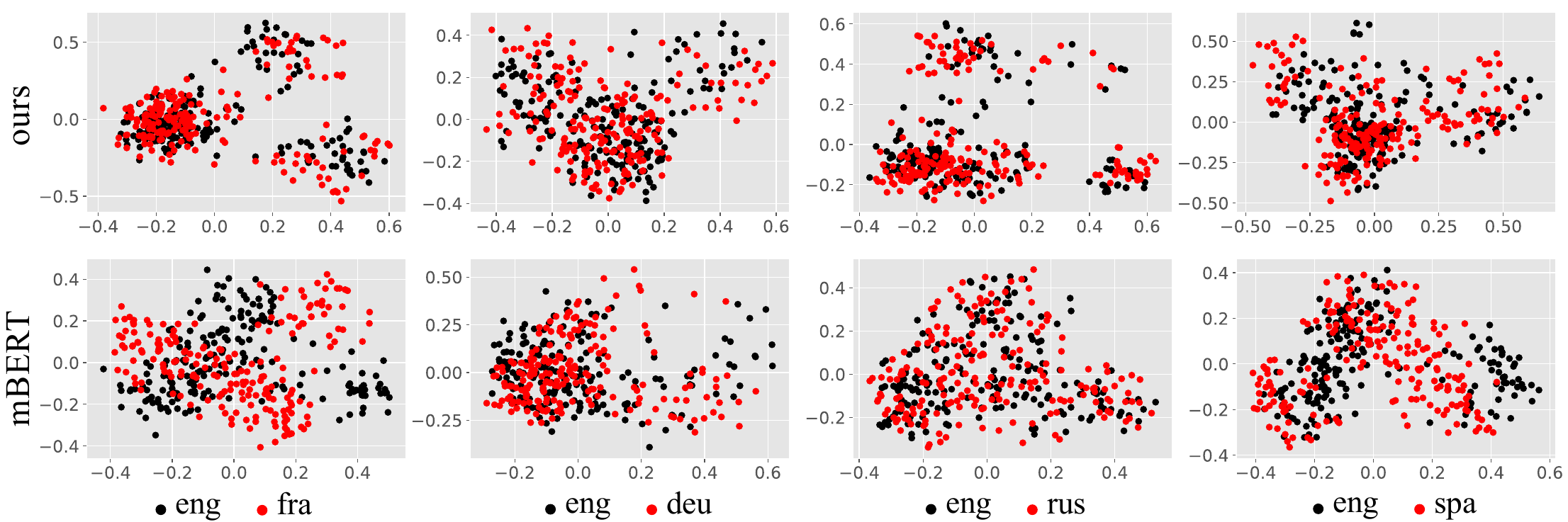}
  \caption{Visualizations of sentence embeddings of CMLM (first row) and mBERT (second row) in Tatoeba dataset in 2D. The target languages are all English and the source languages are French, German, Russian and Spanish.}
  \label{fig:proj}
\end{figure*}

\subsection{Language Agnostic Properties}
\textbf{Language Agnosticism} has been a property of great interest for multilingual representations. However, there has not been a \textbf{qualitative} measurement or rigid definition for this property. We propose that ``language agnostic'' refers to the property that sentences representations are neutral w.r.t their language information. E.g., two sentences with similar semantics should be close in embedding space whether they are of the same languages or not. To capture this intuition, we convert the PAWS-X dataset \citep{paws-x} to a retrieval task to measure the language agnostic property. Specifically, PAWS-X consists of English sentences and their translations in other six languages. Given a query, we inspect the language distribution of the retrieved sentences. The similarity between a query $\vv_{l_1}$ in language $l_1$ and a candidate $\vv_{l_2}$ in language $l_2$ is computed as the cosine similarity $\frac{\vv_{l_1}^T \vv_{l_2}}{\|\vv_{l_1}\|_2\|\vv_{l_2}\|_2}$. In \Cref{fig:la}, representations of mBERT have a strong self language bias, i.e. sentences in the language matching the query are dominant. In contrast, the bias is much weaker in our model, probably due to the cross-lingual retrieval pretraining.


We also discover that removing the first principal component of each monolingual space from sentence representations effectively eliminates the self language bias. Given a monolingual space $\mM^{l_1} \in \R^{N\times d}$, where each row of $\mM^{l_1}$ is a embedding in language $l_1$. For example, in the evaluation on Tatoeba dataset, the monolingual space matrix $\mM^{l_1}$ is computed with texts in language $l_1$ in Tatoeba. The principal component $\vc_{l_1}$ is the first right singular vector of $\mM^{l_1}$. Given a representation $\vv_{l_1}$ in language $l_1$, the projection of $\vv_{l_1}$ onto $\vc_{l_1}$ is removed: $\hat{\vv}_{l_1} = \vv_{l_1} - \frac{\vv_{l_1}^T \vc_{l_1}}{\|\vv_{l_1}\|_2}$. The similarity score between $\vv_{l_1}$ and $\vv_{l_2}$ for cross-lingual retrieval is computed as: $\frac{\hat{\vv}_{l_1}^T \hat{\vv}_{l_2}}{\|\hat{\vv}_{l_1}\|_2\|\hat{\vv}_{l_2}\|_2}$.

As shown in the second and the fourth column in \Cref{fig:la}, with principal component removal (PCR), the language distribution of retrieved texts is much more uniform. We also explore PCR on the Tatoeba dataset. \Cref{tab:tatoeba_pcr} shows the retrieval accuracy of multilingual model with and w/o PCR. PCR increases the overall retrieval performance for both models. This suggests \textbf{the first principal components in each monolingual space primarily encodes language identification information}.

We also visualize sentence embeddings on Tatoeba dataset in \Cref{fig:proj}. Our model shows both weak and strong semantic alignment \citep{lareqa}. Representations are close to others with similar semantics regardless of their languages (strong alignment), especially for French and Russian, where representations form several distinct clusters. Also representations from the same language tend to cluster (weak alignment). While representations from mBERT generally exhibit weak alignment.

\section{Conclusion}
We present a novel sentence representation learning method Conditional Masked Language Modeling (CMLM) for training on large scale unlabeled corpus. CMLM outperforms the previous state-of-the-art English sentence embeddings models, including those trained with (semi-)supervised signals. For multilingual representations, we discover that co-training CMLM with bitext retrieval and cross-lingual NLI finetuning achieves state-of-the-art performance. We also find that multilingual representations have the same language bias and principal component removal can eliminate the bias by separating language identity information from semantics.

\section*{Acknowledgments}
We would like to thank our teammates from Descartes, Google Brain and other Google groups for their feedback and suggestions. We also thank anonymous reviewers for their comments. Special thanks goes to Chen Chen and Hongkun Yu for help with TensorFlow model garden, and Arno Eigenwillig for help on releasing models on TensorFlow Hub.

\bibliography{anthology}
\bibliographystyle{acl_natbib}

\clearpage
\appendix
\section{Methods for Representations}

We evaluate different representations method in Transformer-base models, including CMLM and BERT base (using the model on official Tensorflow Hub). The experiments are conducted on SentEval. Results in \Cref{tab:rep} show that MEAN representation exhibit better performance than CLS and MAX representations.

\begin{table*}[htbp]
\vskip -0.1in
\centering
\resizebox{\textwidth}{!}{\begin{tabular}[c]{l|ccccccccccc}
\toprule
  \textbf{Model} & \textbf{MR}  & \textbf{CR}   & \textbf{SUBJ} & \textbf{MPQA} & \textbf{SST}  & \textbf{TREC} &  \textbf{MRPC} & \textbf{SICK-E} & \textbf{SICK-R} & \textbf{Avg.}\\ \midrule
CMLM MAX & 82.8 & 88.9 & 96.2 & 89.2 & 87.81 & 89.8 & 72.1 & 82.1 & 83.7 & 85.8 \\
CMLM MEAN & 83.6 & 89.9 & 96.2 & 89.3 & 88.5 & 91.0 & 69.7 & 82.3 & 83.4 & \textbf{86.0}  \\
CMLM CLS & 79.1 & 84.3 & 94.2 & 86.9 & 84.9 & 82.6 & 68.4 & 79.3 & 81.7 & 82.4 \\ \midrule
BERT base MAX & 79.6 & 85.5 & 94.6 & 87.3 & 83.0 & 90.0 & 65.6 & 75.5 & 78.1 & 82.1 \\
BERT base MEAN & 81.6 & 87.4 & 95.2 & 87.8 & 85.8 & 90.6 & 71.1 & 79.3 & 80.5 & \textbf{84.3} \\
BERT base CLS & 79.9 & 83.9 & 93.8 & 85.4 & 86.1 & 81.0 & 69.5 & 62.5 & 48.8 & 76.8 \\
\toprule
\end{tabular}}
\vskip -0.1in
\caption{Performance of sentence representations model with different representations method (MAX, MEAN and CLS).}
\vskip -5in
\label{tab:rep}
\end{table*}

\section{Experiments with different Masking ratios}
We test with different masking ratios in CMLM training data. Specifically, We tried masking 40, 60, 80 and 100 tokens of 256 tokens in the CMLM data. Performance of obtained models on SentEval are presented in \Cref{tab:mask}.
\begin{table*}[htbp]
\centering
\resizebox{\textwidth}{!}{\begin{tabular}[c]{c|ccccccccccc}
\toprule
  \textbf{Mask Tokens} & \textbf{MR}  & \textbf{CR}   & \textbf{SUBJ} & \textbf{MPQA} & \textbf{SST}  & \textbf{TREC} &  \textbf{MRPC} & \textbf{SICK-E} & \textbf{SICK-R} & \textbf{Avg.}\\ \midrule
40 & 81.8 & 89.3 & 95.3 & 87.8 & 87.0 & 90.2 & 68.5 & 77.5 & 77.6 & 83.9\\
60 & \textbf{83.7} & 89.5 & 95.8 & 88.9 & 88.0 & 90.3 & 68.7 & 79.5 & 82.8 & 85.4 \\
\textbf{80} & 83.6 & \textbf{89.9} & \textbf{96.2} & \textbf{89.3} & \textbf{88.5} & \textbf{91.0} & 69.7 & \textbf{82.3} & \textbf{83.4} & \textbf{86.0} \\
100 & 83.2 & 89.5 & 95.5 & 88.7 & 88.0 & 90.8 & \textbf{70.0} & 81.5 & 82.7 & 85.6 \\
\toprule
\end{tabular}}
\label{tab:mask}
\vskip -0.1in
\caption{Performance with different masking ratios in data (X-out-of-256) of CMLM base on SentEval.}
\vskip -5in
\end{table*}

\section{Training Configurations and Implementation Details}

\textbf{Projection $P$ in the CMLM modeling.} Let h denote the dimension of the input sentence vector (e.g. h = 768 in BERT base; h = 1024 in BERT large). Let $FC (h_1, h_2, n)$ denote a fully connected layer with input dimension $h_1$, output dimension $h_2$ and nonlinearity function $n$. The three layers are $FC(h, 2\times h, \text{ReLU})$, $FC(2\times h, 2\times h, \text{ReLU})$, $FC(2\times h, h, \text{None})$. We tried projections without intermediate layers and observed a drop in training LM accuracy. Adding more layers doesn’t improve the MLM accuracy or downstream tasks performance. Using $2\times h$ is empirically chosen based on preliminary experiments. Other hidden sizes are also explored.

\textbf{Configurations for multilingual representations learning.} In general, larger batch sizes improve performance until we reach ~2048, since each example will see more “mismatched” examples. After 2048, we do not see obvious improvements in performance from increasing batch size. We’ll add detailed results on this in the final version. The training steps for different stages are decided on a validation set.

\textbf{Training Data and infrastructure.} English pretraining takes 5 days on 64 TPUs using 1TB of data from Common Crawl dumps 2020-1, 2020-05, 2020-10. More data could be beneficial, but would increase training time.

\section{Reliability of XEVAL}
In this section, we want to discuss the reliability of XEVAL. XEVAL contains sentence-level data and we expect its translation not to be too challenging. Inspection by in-house bilingual speakers also confirms the high quality of translation. Human translation is always preferred but we are limited by budget and annotator resources (especially for low-resource languages).

\section{CMLM's Comparison with Next Sentence Prediction (NSP) and Potential Limitations.}
We tried MLM (CC) with and w/o NSP and it does not make much difference on SentEval. Training NSP accuracy quickly converge to 95\%, indicating that NSP is not a challenging task.

Sentence embedding methods like CMLM can be less effective for sequence labeling (e.g., NER) and natural language generation (NLG) and question answering (Q\&A).

\section{Performance Variances}
We provide the performance variances of CMLM base and CMLM large on SentEval dataset in \Cref{tab:var}.
\begin{table*}[htb]
\small
\centering
\resizebox{\textwidth}{!}{\begin{tabular}[c]{l|cccccccccccc}
\toprule
  \textbf{Model} & \textbf{MR}  & \textbf{CR}   & \textbf{SUBJ} & \textbf{MPQA} & \textbf{SST}  & \textbf{TREC} &  \textbf{MRPC} & \textbf{SICK-E} & \textbf{SICK-R}\\ \midrule
CMLM base & 83.6$\pm$0.2 & 89.9$\pm$0.4 & 96.2$\pm$0.1 & 89.3$\pm$0.3 & 88.5$\pm$0.2 & 91.0$\pm$0.8 & 69.7$\pm$0.6 & 82.3$\pm$0.3 & 83.4$\pm$0.4 \\
CMLM large & 85.6$\pm$0.2 & 89.1$\pm$0.3 & 96.6$\pm$0.2 & 89.3$\pm$0.3 & 91.4$\pm$0.1 & 92.4$\pm$0.7 & 70.0$\pm$1.0 & 82.2$\pm$0.5 & 84.5$\pm$0.4 \\ \toprule
\end{tabular}}
\vskip -0.1in
\caption{Performance variances of CMLM base and CMLM large on SentEval.}
\label{tab:var}
\end{table*}

\end{document}